\documentclass{article}
\usepackage{spconf,amsmath,graphicx}
\usepackage{float}
\usepackage{algorithmic}
\usepackage[ruled,vlined]{algorithm2e}
\usepackage{threeparttable}
\usepackage{booktabs}
\usepackage{tabulary}
\usepackage{multirow}
\usepackage{multicol}
\usepackage{color,soul}
\usepackage{subfigure}

\title{Quantile Online Learning for Semiconductor Failure Analysis}
\name{Bangjian Zhou$^{\star}$, Pan Jieming$^{\dagger}$, Maheswari Sivan$^{\dagger}$, Aaron Voon-Yew Thean$^{\dagger}$, J. Senthilnath$^{\star}$\thanks{This study is supported by the Accelerated Materials Development for Manufacturing Program at A*STAR via the AME Programmatic Fund by the Agency for Science, Technology and Research under Grant No. A1898b0043.}}
\address{$^{\star}$Institute for Infocomm Research (I$^{2}$R), A*STAR, Singapore\\
$^{\dagger}$Electrical and Computer Engineering, National University of Singapore, Singapore}

\begin{document}
%
\maketitle
\begin{abstract}
With high device integration density and evolving sophisticated device structures in semiconductor chips, detecting defects becomes elusive and complex. Conventionally, machine learning (ML)-guided failure analysis is performed with offline batch mode training. However, the occurrence of new types of failures or changes in the data distribution demands retraining the model. During the manufacturing process, detecting defects in a single-pass online fashion is more challenging and favoured. This paper focuses on novel quantile online learning for semiconductor failure analysis. The proposed method is applied to semiconductor device-level defects: FinFET bridge defect, GAA-FET bridge defect, GAA-FET dislocation defect, and a public database: SECOM. From the obtained results, we observed that the proposed method is able to perform better than the existing methods. Our proposed method achieved an overall accuracy of $86.66\%$ and compared with the second-best existing method it improves $15.50\%$ on the GAA-FET dislocation defect dataset.
\end{abstract}
\section{Introduction}
\label{sec:intro}
For several decades the semiconductor industry follows Moore's law to drive development, and billions of well-designed semiconductor chips are put into use every year. Fine chips require more care about defects. Notably, as the devices are subjected to aggressive channel length scaling, the impact of bridge and dislocation defects on the transport properties of the transistor becomes aggravated. As a result, engineering efforts are essential for semiconductor failure analysis (FA) to detect defects on time and guarantee major components work rightly \cite{park2020review}.

Lately, Machine Learning (ML) has become increasingly important to accelerate semiconductor FA. There are mainly 2 kinds of the semiconductor dataset used in the ML model to learn and generalize, (1) Unstructured dataset, for example, wafer map imagery used to understand the wafer quality and identify the defects patterns \cite{chien2020inspection} and (2) Structured dataset, for example, the transfer characteristics (drain current ($I_d$) vs. gate voltage ($V_g$)) of the transistor for different position of bridge defects \cite{ACS1}. The structured dataset is favoured by researchers as the features like slope and intercept (for example, sub-threshold swing and threshold voltage for transistor) represent healthy/unhealthy states more discriminatively. The structured datasets' features differ a lot in ranges, which is attributed to the operational principle of a transistor as a logic switch where the drain current varies across $\sim$8 orders of magnitude, whereas drain voltage varies from 0-1V (for N-channel Metal-oxide Semiconductor).

The ML models used for the FA can be divided as traditional ML, such as tree-based model \cite{ACS1}, Neuro-fuzzy system (NFS) \cite{ferdaus2022}, and Deep Neural Network (DNN) model \cite{chien2020inspection}. The DNN models are affected by the high-range features that would take a dominant place, making other features insignificant. Hence, data normalization is necessary for DNN training. Meanwhile, as the number of semiconductor chips is increasing exponentially, the scenario has changed. The offline batch-mode DNN training fails to keep up pace with the large data stream, online single-pass mode DNN training is preferred. This makes the data normalization for the DNN difficult. Though we have many well-designed methods like min-max scaling, z-score normalization \cite{patro2015normalization} and quantile normalization \cite{zhao2020quantile} for offline training, whereas in the online scenario, we lack global statistics requested by the normalization methods. Although some researchers have proposed using sliding windows to get statistics or designing equations/layers to approximate the statistics used in the normalization, most online normalization methods are still designed in batch-mode, whereas single-pass mode is rarely adopted.

In this paper, we propose a novel quantile online learning method for semiconductor FA. The main contributions of this study are summarized as follows:
\begin{figure}[tb]
\begin{minipage}[hbtp]{1.0\linewidth}
  \centering
\centerline{\includegraphics[height=5cm,width=8.5cm]{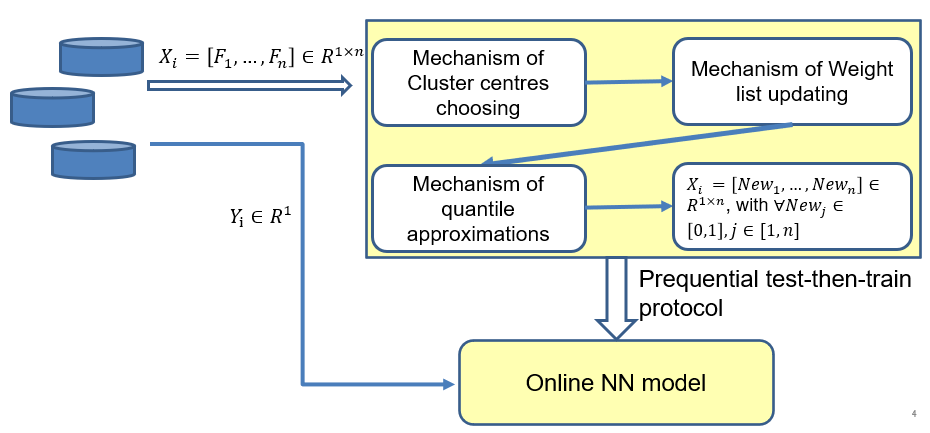}
}
\end{minipage}
\caption{\centering The work flow of the quantile online learning method}
\label{fig:workflow}
\vspace{-0.15in}
\end{figure}
(1) Unlike quantile normalization averaging the feature values in batch/mini-batch setting with the same quantile value assigned for different samples, we use quantile numbers to replace raw features. Here, we observe the proposed approach is suitable for semiconductor datasets with varying feature ranges.
(2) We trained the online DNNs in a purely single-pass mode including the input data normalization, there are very few methods performing semiconductor FA in such a way.
(3) We propose a discount factor to control the proportion of local or global information in the statistics, which helps with defect detection. 
(4) Our experimental results demonstrate that the proposed method perform better than the existing methods on four semiconductor defect detection datasets, which exceeds the second-best existing method by $15.50\%$.

\section{Related work}
\label{sec:related_work}
\textbf{ML-guided FA.} Many traditional ML-based approaches have been applied to semiconductor FA. He et al., \cite{4369338} utilized a k-nearest neighbor (KNN) rule-based algorithm to handle faults. Xie et al., \cite{xie2014novel} applied a support vector machine to detect defect patterns in the image. Nawaz et al., \cite{nawaz2014fault} developed a Bayesian network for etching equipment fault detection. After Lecun et al., \cite{lecun2015deep} pointed out that DNN can improve the state-of-the-art methods in many fields, researchers also tried DNN methods, while the traditional methods are still common. Hsu et al., \cite{hsu2021multiple}, Kim et al., \cite{kim2019fault} and Chien et al., \cite{chien2020inspection} proposed some variants of convolutional neural network for FA in image datasets. Pan et al., proposed a tree-based transfer learning model from FinFET to GAA-FET \cite{ACS1}. Kim et al., \cite{kim2021variational} devised a generative model to handle process drift in the defect detection scenario. Ferdaus et al., \cite{ferdaus2022} devised an online NFS for semiconductor FA. However, most methods are focused on offline settings, this motivates us to design an efficient online learning methods for DNNs.

\noindent\textbf{Input data normalization.} Commonly used offline input data normalization strategies are min-max scaling, z-score normalization and decimal scaling \cite{patro2015normalization}. However, these techniques need to adapt to online scenarios and adaptive versions are few. To the best of our knowledge, the adaptive normalization methods can be divided into two categories: (1) Ogasawara et al., \cite{ogasawara2010adaptive}, Gupta et al., \cite{gupta2019adaptive} used sliding window to adapt statistics in streaming data. (2) Passalis et al., \cite{passalis2019deep} proposed DAIN that uses additional linear layers to mimic the adaptive mean and standard deviation value for normalization. Tran et al., \cite{tran2021data} and Shabani et al., \cite{shabani2022multi} proposed improved variants of DAIN to handle time-series data. Though these methods were devised for online settings, they are designed in batch-mode. Our proposed method is a single-pass design, which means there is hardly any latency in getting predictions. And quantile normalization is another extremely popular method that could produce well-aligned distributions so all samples are from the same distribution, which is commonly seen in gene expression dataset \cite{hansen2012removing}, \cite{zhao2020quantile}. Note, our proposed method is totally different from quantile normalization; as they would not use the quantile number to replace the feature value.

\section{Quantile online learning}
\label{sec:methodology}
The proposed quantile online learning (QOL) classifies semiconductor FA datasets on the fly. The flow diagram of the proposed QOL is shown in Fig. \ref{fig:workflow} and discussed in detail in the below subsections.

\subsection{Problem statement}
\label{sec:problem_statement}
The semiconductor structured input data stream i.e, $X_i=[F_1,...,F_n] \in \Re^{1 \times n}$, $i\in \mathcal{N}$, which means $i^{th}$ sample and $n$ is the number of dimensions. Each input has a corresponding label $Y_i \in \mathcal{N}$ that represents the defect types; our objective is to address the classification problem of the defect types. To mimic real-world settings, a prequential test-then-train protocol is applied. This protocol would use the model to give predictions and then do the training.

\subsection{Buffer list initialization}
\label{sec:buffer_list_initialization}
Initially, a small list of samples is collected i.e, $C=[X_1,...,\\ X_p]\in \Re ^{p\times n}$, where $p$ is the number of samples, set between 10 to 30. We collect the feature values in a 2D buffer list from samples $C$ i.e, 
V = $[[V_{11},..,V_{1p}],$
$..[V_{n1},..,V_{np}]]$. The $V_{ij}$ means the feature value $F_i$ from sample $X_j$. Next, we sort each single 1D list independently and define another 2D weight list $W=[[1,..,1],..,[1,..,1]]$ with all ones. We treat the buffer list $V=\{V_{ij}\}^{n\times p}$ as cluster centres which capture the knowledge of the distribution of raw data, and weight list $W= \{W_{ij}\}^{n\times p}$ as the number of samples in the cluster. During this phase, we wait for $p$ samples to start, whereas the remaining steps use single-pass mode.

\subsection{Cluster centre choosing}
\label{sec:cluster_centre_choosing}
We obtain the initial $V$ and $W$ and the raw input features $X_i$ from the data stream. We choose the cluster centres for each feature by calculating distance $dis(V_{ij},F_i)=||V_{ij}-F_i||_2=\sqrt{(V_{ij}-F_i)^2}$ and find the index of minimum distance for the $k^{th}$ feature using $$idx_k=\{i|\min{dis(V_{ki},F_k)},1\leq i \leq p \},$$ 
The index list $idx =[idx_1,...,idx_n]$ is used as a reference to update later steps.

\subsection{Weight list updating}
\label{sec:weight_list_updating}
\begin{figure*}[{t}]
\subfigure[case(i) idea]{
\begin{minipage}[hbtp]{0.5\linewidth}
  \centering
\centerline{\includegraphics[height=1.7cm,width=7.2cm]{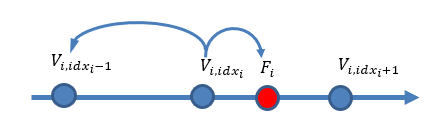}}\medskip
\label{fig:case1}
\end{minipage}
}
\subfigure[case(ii) idea]{
\begin{minipage}[hbtp]{0.5\linewidth}
  \centering
\centerline{\includegraphics[height=1.7cm,width=7.2cm]{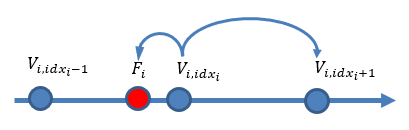}}\medskip
\label{fig:case2}
\end{minipage}
}
\vspace{-0.15in}
\caption{The general ideas for weight list updating}
\end{figure*}

The main idea behind updating is to replace the old cluster centre with the new feature value and make the corresponding weight contribute to the closest cluster centres. We consider 2 cases in updating: (i) $F_i$ $\geq$ $V_{i,idx_i}$ and (ii) $F_i$ $<$ $V_{i,idx_i}$. For case (i) in Fig \ref{fig:case1}, $old\_weight = W_{i,idx_i}$ represents the weight for $V_{i,idx_i}$ before updating. We first calculate a $percentage$ factor as below:
\begin{align}
percentage = \frac{dis(V_{i,idx_i},F_i)}{dis(V_{i,idx_i-1},F_i)}
\label{eq:percentage1}
\end{align}
Then we update $W_{i,idx_i}, W_{i,idx_i-1}$ using,
\begin{align}
  W_{i,idx_i} &= (1-percentage)\times old\_weight+1
  \label{eq:Widx1}\\
  W_{i,idx_i-1} &= W_{i,idx_i-1}+percentage\times old\_weight
  \label{eq:Widx2}
\end{align}
These equations make sure the closer the value is to the cluster centres, the more weightage is assigned from the old weight. If the $W_{i,idx_i-1}$ does not exist, we give all weight to the new cluster using,
\begin{align}
  W_{i,idx_i} &= old\_weight+1
  \label{eq:Widx3}
\end{align}
For case (ii) in Fig.\ref{fig:case2}, we still use $old\_weight$ as above, and only give modifications to the above equations as follows:
\begin{align}
percentage &= \frac{dis(V_{i,idx_i},F_i)}{dis(V_{i,idx_i+1},F_i)}
\label{eq:percentage2}\\
  W_{i,idx_i} &= (1-percentage)\times old\_weight+1
  \label{eq:Widx4}\\
  W_{i,idx_i+1} &= W_{i,idx_i+1}+percentage\times old\_weight
  \label{eq:Widx5}
\end{align}
For the specific case $W_{i,idx_i+1}$ does not exist, then we use Eq.(\ref{eq:Widx3}) to update. Next, we introduce a discount factor $\eta$, which is used to control the forget rate. To make the buffer list capture a sense of local trend, we set $\eta$ between 0 to 1. After each sample's all features modification, we would update the whole weight list $\{W_{ij}\}^{n\times p}=\eta \times \{W_{ij}\}^{n\times p}$. Finally, we assign $V_{i,idx_i}=F_i$ as the new cluster centre.

\subsection{Quantile approximations}
\label{sec:quantile_approximations}
After we find the suitable position for the current sample's feature values, we replace all the feature values $F_i$ using,
\begin{align}
F_i &=  \sum\limits_{j=1}^{idx_i}{W_{ij}}/\sum\limits_{j=1}^{p}{W_{ij}}
\label{eq:calc_quantile}
\end{align}
The weight for each cluster centres is used to approximate the number of feature values that are less than it so that we obtain an approximation of the quantile for each feature value. Then we feed the normalized data $X_i=[New_1,...,New_n]$ to train the online DNN model.

\begin{algorithm}
  \KwIn{sample from data stream $\mathbf{X_i}$ with label $Y_i$, buffer list length $p$, discount factor $\eta$}
  \KwOut{Normalized data stream $\mathbf{\Hat{X_i
  }}$, trained online NN model $\mathbf{H}$, predictions $\Hat{Y}_i$ of sample $X_i$}
  Initialize the buffer list $V$ and weight list $W$ (sec \ref{sec:buffer_list_initialization})\par
  \While{$X_i$ is not None}{
    \For{$j=1~ \text{to}~ n$}
    {calculate the minimum distance between $F_j$ and $[V_{j1},...,V_{jp}]$, get index $idx_j$ (sec \ref{sec:cluster_centre_choosing})\par
    \If{$F_j \geq V_{j,idx_j}$}{Use Eqs.(\ref{eq:percentage1}-\ref{eq:Widx3}) to update $\mathbf{\{W_{jk}\}^{1\times p}}$,$1 \leq k \leq p$}
    \If{$F_j$ \textless $V_{j,idx_j}$}
    {Use Eqs.(\ref{eq:Widx3}-\ref{eq:Widx5}) to update $\mathbf{\{W_{jk}\}^{1\times p}}$,$1 \leq k \leq p$}
    Calculate $newF_j$ with Eq.(\ref{eq:calc_quantile}),\par
    $\mathbf{\Hat{X_i
  }[j] = newF_j}$
    }
    $W = \eta \times \{W_{ij}\}^{n\times p}$ \par
    Train the NN model with the  test-then-train protocol using normalized data $\mathbf{\Hat{X_i
  }}$
  }
  \caption{Quantile online learning}
\end{algorithm}

\section{EXPERIMENTAL RESULTS}
\label{sec:experimental_results}
\textbf{Data description.}
In this work, three real-world multi-class semiconductor fault detection datasets and a public dataset SECOM \cite{secomuci} from the UCI machine learning repository are considered. The real-world datasets include bridge defects in GAA-FET (GAA-FET-b), dislocation defects in GAA-FET (GAA-FET-d) and bridge defects in FinFET (FinFET-b). These FETs datasets were generated using a full three-dimensional technology computer aided design (TCAD) digital twin model and then were calibrated with the Current-Voltage curve in the actual device defects \cite{ACS1}. The SECOM dataset consists of complex signals for the engineers to analyze the defects. The detailed information of each dataset is shown in Table \ref{table:dataset}. Classifying imbalance classes in the dataset is more challenging as the minority class would only be $3\sim20\%$ of the majority class, say, SECOM is 104 fails to 1463 pass, GAA-FET-b is 27 to 764 and FinFET-b is 79 to 512. As a result, overall accuracy and class balanced accuracy were used as evaluation metrics.   
\begin{table}[t]
\centering
\caption{The detailed information about datasets}
\label{table:dataset}
\begin{tabular}{llll}
\hline
Dataset & dimension & classes & number of samples \\
\hline
GAA-FET-b  & 23 & 12 &3051    \\
GAA-FET-d & 40 & 9 & 90 \\
FinFET-b   & 23 & 10 &2739\\
SECOM   & 590 & 2 & 1567 \\
\hline
\end{tabular}
\vspace{-0.15in}
\end{table}

\noindent\textbf{Experiment setting.} We chose the Adaptive Normalization (AN) from \cite{gupta2019adaptive} and Deep Adaptive Input Normalization (DAIN) from \cite{passalis2019deep} as baseline methods to compare the data normalization, and passing the raw data. As the baseline methods are variants of min-max and z-score normalization, which cannot handle feature values with outliers pretty well. Hence, we additionally pre-processed the data by logarithm transformation to make the feature range small, which is not adapted for our proposed method. We chose three online DNN models, which include a DNN with 2 layers fully connected feed-forward network, Deep Evolving Denoising Autoencoder (DEVDAN) from \cite{ashfahani2020devdan} and Autonomous Deep Learning (ADL) model from \cite{ashfahani2019autonomous} to classify the normalized data in a single-pass mode. The ADL and DEVDAN are two evolving architectures designed for online settings, which would autonomously generate the needed hidden nodes. To make a fair comparison, we limited different models to all have 2 layers and the nearly same amount of parameters. We intended to prove our method's robustness by using these models. We applied the prequential test-then-train protocol to the whole datasets, except for GAA-FET-d as current online learning models rarely get good performance in limited sequences (90 samples). We split and augment the train set and evaluate the test set with the test-then-train protocol.  

We fine-tuned the hyper-parameters in the following range: [0.1,0.5] as $lr_{disc}$ and [$1e^{-3}$,$1e^{-1}$] as $lr_{gen}$ for DEVDAN \cite{ashfahani2020devdan}; [0.05,0.35] as $lr$ for ADL \cite{ashfahani2019autonomous}; [$1e^{-4}$,$1e^{-1}$] as $lr$ for DNN; batch size 2 to 5 and threshold $\delta$ 0.1 to 0.3 for AN \cite{gupta2019adaptive}; $mean\_lr$ in [$1e^{-8}$,$1e^{-1}$], $scale\_lr$ in [$1e^{-8}$,$1e^{-1}$], $gate\_lr$ in [$1e^{-3}$,$1$] and batch size 5 for DAIN \cite{passalis2019deep}. We select the batch size to be small so that it's a relatively fair comparison with our single-pass method. 

\begin{table}
\begin{threeparttable}
\caption{Classification accuracy in semiconductors FA }\label{tab:overall_accu}
\begin{centering}
{\begin{tabular}{>{\raggedright}p{0.8cm}>{\raggedright}p{0.9cm}>{\centering}p{1.6cm}>{\centering}p{1.6cm}>{\centering}p{1.6cm}}
\hline 
\multirow{2}{1cm}{Dataset} & \multirow{2}{1cm}{pre- \\process} & \multicolumn{3}{c}{Overall accuracy/balanced accuracy}\tabularnewline
\cline{3-5} 
 &  & DNN & ADL &  DEVDAN\tabularnewline
\hline 
\multirow{4}{2cm}{GAA- \\FET-b} & Raw &81.61/62.21  & 79.06/48.96 &  78.72/66.86\tabularnewline
 & AN & 74.73/44.45 & 87.00/66.70 & \textbf{86.95}/66.32\tabularnewline
 & DAIN & 62.66/38.39 & 74.56/56.34 & 74.68/54.28 \tabularnewline
 & QOL & 79.15/53.29 & 85.61/\textbf{67.73} &  85.80/69.26\tabularnewline
\hline 
\multirow{4}{2cm}{GAA- \\FET-d} & Raw & 51.11/57.03 & 42.22/45.18 &  40.00/43.70\tabularnewline
 & AN & 53.33/55.18 & 54.66/60.00 & 49.99/55.92\tabularnewline
 & DAIN & 43.33/40.00 & 32.96/32.22 & 43.33/45.18 \tabularnewline
 & QOL & \textbf{98.88}/\textbf{99.26} & \textbf{98.88}/\textbf{99.26} &  97.77/98.52\tabularnewline
\hline 
\multirow{4}{2cm}{Fin- \\FET-b} & Raw &70.85/63.02  & 71.55/70.53 &  83.64/80.69\tabularnewline
 & AN & 72.05/61.44 & 91.86/87.27 & 92.02/87.65\tabularnewline
 & DAIN & 58.40/46.49 & 83.15/81.05 & 82.86/80.59 \tabularnewline
 & QOL & 76.01/68.63 & \textbf{92.56}/88.74 &  92.83/\textbf{88.92}\tabularnewline
\hline 
\multirow{4}{2cm}{SE-\\COM} & Raw & 92.53/51.34 & \textbf{92.26}/51.11 & 92.47/51.22  \tabularnewline
 & AN & 90.72/52.87 & 90.81/52.92 & 92.29/51.66\tabularnewline
 & DAIN & 91.90/51.18 & 90.55/53.32 & 90.55/52.60 \tabularnewline
 & QOL & 92.44/51.65 & 90.78/\textbf{53.53} &  92.17/52.22\tabularnewline
\hline 
\end{tabular}}
\par\end{centering}
\end{threeparttable}
\label{table:performance}
\end{table}

\noindent\textbf{Performance comparisons.} We compare the performance of 3 data normalization approaches (AN, DAIN, QOL) and Raw data on semiconductor datasets that are combined with 3 DNN models separately. For the 3 multi-class semiconductor datasets, from Table \ref{tab:overall_accu}, we observe that our QOL performs well for both overall and balanced accuracy while the AN was the second best, and even got better overall accuracy in GAA-FET-b. However, the DAIN method has low performance, due to (1) the data sequence being limited, and DAIN's converge rate being affected by more network parameters. (2) a small batch size may not be suitable for this method. Compared with the Raw data, AN and QOL's result is a strong demonstration of the need for data normalization. For the SECOM, we emphasize the balanced accuracy here as most methods can get high overall accuracy by predicting all as the majority. Our proposed QOL achieves the best balanced accuracy, while the DAIN, as well as AN, also performed better for this high-dimensional complex dataset. All the normalization methods can address the imbalance problems a bit, which may be due to normalized features being more discriminative. Among all the datasets, we find that our single-pass QOL is more stable in different settings and gets comparable or better performance than these batch-mode normalization methods.

\section{CONCLUSIONS}
\label{sec:conclusions}
In this paper, we proposed a novel quantile online learning approach to address single-pass online classification for semiconductor FA. QOL utilizes the single-pass online framework where quantile representation is passed to DNN architecture to adapt to the change in the feature distribution of different semiconductor defects by prequential test-then-train protocol. This helps in achieving better classification performance in both overall accuracy and balanced accuracy. We compared the performance of our technique with the existing methods for a binary class and three multi-class semiconductor defect datasets, for all these datasets the proposed method performed better in comparison with the existing methods.

\newpage

\bibliographystyle{IEEEbib}
\bibliography{Brefs}

\end{document}